\pdfoutput=1
\documentclass[11pt,a4paper]{article}
\usepackage[hyperref]{acl2021}
\usepackage{times}
\usepackage{latexsym}

\usepackage{multirow}
\usepackage{inputenc}
\usepackage{multirow}
\usepackage{flushend}
\usepackage{times}
\usepackage{enumitem}
\usepackage[running]{lineno}
\usepackage{flushend}
\usepackage{todonotes}
\usepackage{tcolorbox}
\usepackage{color,soul}
\usepackage[section]{placeins}
\usepackage{epstopdf}
\usepackage{multirow,tabularx}
\usepackage{mdframed} 
\usepackage[multiple]{footmisc}
\usepackage{enumitem}
\usepackage{arevmath}    
\usepackage{microtype}
\aclfinalcopy

\title{Quality Evaluation of the Low-Resource Synthetically Generated Code-Mixed Hinglish Text}

\author{Vivek Srivastava \\
  TCS Research\\ Pune, Maharashtra, India \\
  \texttt{srivastava.vivek2@tcs.com} \\\And
  Mayank Singh \\
  IIT Gandhinagar\\ Gandhinagar, Gujarat, India \\
  \texttt{singh.mayank@iitgn.ac.in} \\}

\date{}

\begin{document}
\maketitle
\begin{abstract}
In this shared task, we seek the participating teams to investigate the factors influencing the quality of the code-mixed text generation systems. We synthetically generate code-mixed Hinglish sentences using two distinct approaches and employ human annotators to rate the generation quality. We propose two subtasks, \textit{quality rating prediction} and \textit{annotators' disagreement prediction} of the synthetic Hinglish dataset. The proposed subtasks will put forward the reasoning and explanation of the factors influencing the quality and human perception of the code-mixed text.    
\end{abstract}

\section{Introduction}
\label{sec: introduction}
Code-mixing is the phenomenon of mixing words and phrases from multiple languages in a single utterance of a text or speech. Figure \ref{fig: example} shows the example code-mixed Hinglish sentences generated from the corresponding parallel Hindi and English sentences. Code-mixed languages are prevalent amongst multilingual communities such as Spain, India, and China. With the inflation of social-media platforms in these communities, the availability of code-mixed data is seeking a boom. It has lead to several interesting research avenues for problems in computational linguistics such as language identification \cite{singh2018language, shekhar2020language}, machine translation \cite{dhar2018enabling, srivastava2020phinc}, language modeling \cite{pratapa2018language}, etc.

\begin{figure}[!tbh]
\centering
\small{
\begin{tcolorbox}[colback=white]
\begin{center}
    \textbf{Example I}
\end{center}
\textsc{Hinglish}: \textcolor{orange}{ye ek} \textcolor{blue}{code mixed sentence} \textcolor{orange}{ka} \textcolor{blue}{example} \textcolor{orange}{hai} \\
\textsc{English }: \textcolor{blue}{this is an example code-mixed sentence} 

\begin{center}
    \textbf{Example II}
\end{center}
\textsc{Hinglish} : \textcolor{orange}{kal me} \textcolor{blue}{movie} \textcolor{orange}{dekhne ja raha hu}. \textcolor{blue}{How are the reviews}?\\
\textsc{English}:
\textcolor{blue}{I am going to watch the movie tomorrow. How are the reviews?}

\end{tcolorbox}}
\caption{Example parallel Hinglish and English sentences. The code-mixed Hinglish sentences contain words from \textcolor{orange}{Hindi} and \textcolor{blue}{English} languages.}
\label{fig: example}
\end{figure}

Over the years, we observe various computational linguistic conferences and workshops organizing the shared tasks involving the code-mixed languages. Diverse set of problems have been hosted such as sentiment analysis \cite{chakravarthi-etal-2021-findings-shared-task, patwa2020semeval}, offensive language identification \cite{chakravarthi-etal-2021-findings-shared-task}, word-level language identification \cite{solorio2014overview, molina2019overview}, information retrieval \cite{banerjee2016overview}, etc.

Despite these overwhelming attempts, the natural language generation (NLG) and evaluation of the code-mixed data remain understudied. The noisy and informal nature of the code-mixed text adds to the complexity of solving and evaluating the various NLG tasks such as summarization and machine translation. These inherent challenges \cite{srivastava2020phinc} with the code-mixed data makes the widely popular evaluation metrics like BLEU and WER obsolete. Various metrics (e.g., CMI \cite{das2014identifying, gamback2016comparing}, M-index \cite{doi:10.1177/13670069000040020101}, I-index \cite{guzman2017metrics}, Burstiness \cite{goh2008burstiness}, Memory \cite{goh2008burstiness}, etc.) have been proposed to measure the complexity of code-mixed data, but they fail to capture the linguistic diversity which leads to poorly estimating the quality of code-mixed text \cite{srivastava2021challenges}. 

With this shared task\footnote{\url{https://sites.google.com/view/hinglisheval}} (see Section \ref{sec: overview} and \ref{sec: task} for the detailed description), we look forward to the new strategies that cater to the broad requirement of the quality evaluation of the generated code-mixed text. These methods will entail various linguistic features encompassing syntax and semantics and the perspectives of human cognition such as writing style, emotion, sentiment, language, and preference. We also put forward a subtask to understand the factors influencing the human disagreement on the quality rating of the generated code-mixed text. This could help design a more robust quality evaluation system for the code-mixed data.

\section{Task Overview}
\label{sec: overview}
In this shared task, we propose two subtasks evaluating the quality of the code-mixed Hinglish text. First, we propose to predict the quality of Hinglish text on a scale of 1--10. We aim to identify the factors influencing the text's quality, which will help build high-quality code-mixed text generation systems. We synthetically generate the Hinglish sentences using two different approaches (see Section \ref{sec: dataset}) leveraging popular English-Hindi parallel corpus. Besides, we also have at least two human-generated Hinglish sentences corresponding to each parallel sentence. The second subtask aims to predict the disagreement on a scale of 0--9 between the two annotators who have annotated the synthetically generated Hinglish sentences. Various factors influence this human disagreement, and we seek to investigate the reasoning behind this behavior.

\section{Dataset}
\label{sec: dataset}
As outlined in Section \ref{sec: introduction}, the code-mixed NLG task observes a scarcity of high-quality datasets. Consequently, the quality evaluation of the generated code-mixed text remains unexplored. We propose a new dataset with Hinglish sentences generated synthetically and rated by human annotators to address this challenge. We create the dataset in two phases.\\
\textbf{Phase 1: Human-generated Hinglish sentences}: We select the English-Hindi parallel sentences from the IIT-B parallel corpus
\cite{kunchukuttan2018iit} to generate the Hinglish sentences. The parallel corpus has 1,561,840 sentence pairs. We randomly select 5,000 sentence pairs, in which the number of tokens in both the sentences is more than five. We employ five human annotators and assign each 1,000 sentence pairs. Table \ref{tab:guidelines} shows the annotation guidelines to generate the Hinglish sentences. Post annotation, we obtain 1,976 sentence pairs for which the annotators have generated at least two Hinglish sentences.

\begin{figure*}[t]
\centering
\begin{tabular}{cc}
    \includegraphics[width=0.4\linewidth]{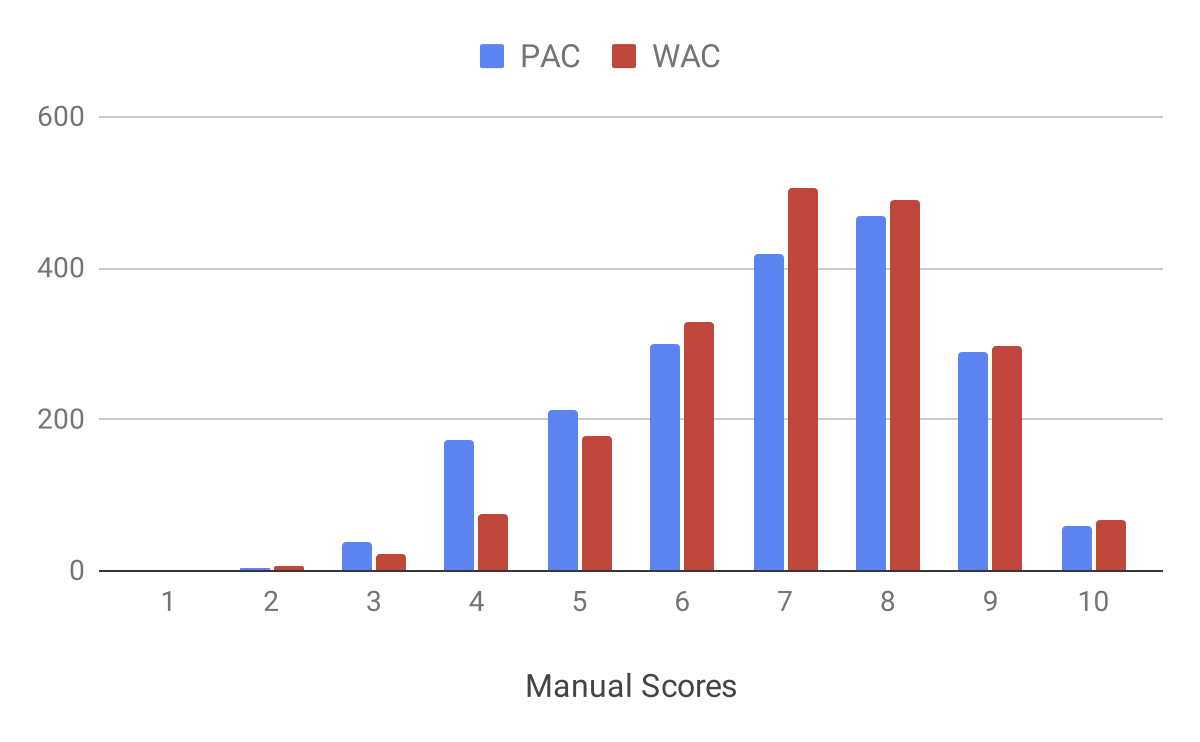} &  \includegraphics[width=0.4\linewidth]{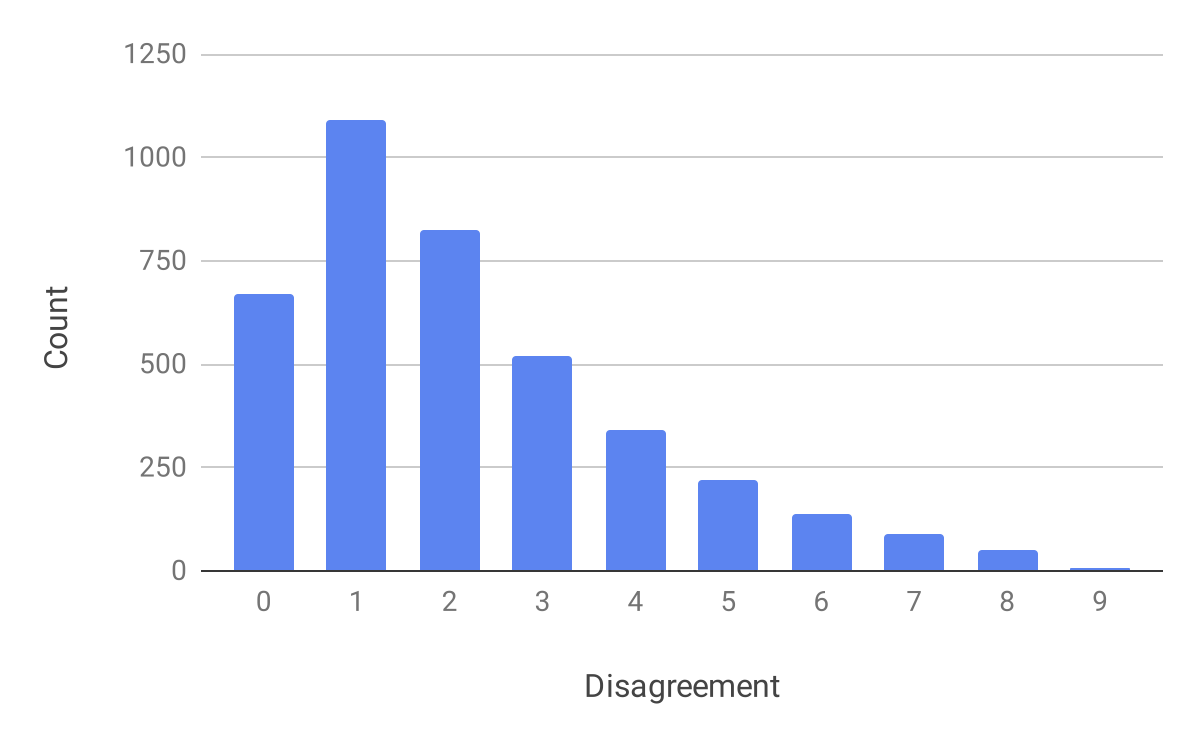}\\
    (a) & (b)\\ 
\end{tabular}

\caption{Distribution of (a) human evaluation scores and (b) disagreement in human scores in the synthetically generated Hinglish sentences.}
\label{fig:dataset_stat}
\end{figure*}

\noindent\textbf{Phase 2: Synthetic Hinglish sentence generation and quality evaluation}:
We synthetically generate the Hinglish sentence corresponding to each of the parallel 1,976 English-Hindi sentence pairs. We employ two different code-mixed text generation (CMTG) techniques:
\begin{itemize}[noitemsep,nolistsep,leftmargin=*]
  \item Word-aligned CMTG (WAC): Here, we align the noun and adjective tokens between the parallel sentences. We replace the aligned Hindi token with the corresponding English token and transliterate the Hindi sentence to the Roman script.  
  \item Phrase-aligned CMTG (PAC): Here, we align the key-phrases of length up to three tokens between the parallel sentences. We replace the aligned Hindi phrase with the corresponding English phrase and transliterate the Hindi sentence to the Roman script.
\end{itemize}
For the token alignment between parallel sentences, we use the online curated dictionaries, GIZA++~\cite{och03:asc} trained on the remaining IIT-B corpus, and cross-lingual word embedding trained on English and Hindi word vectors from FastText~\cite{bojanowski2017enriching}. We employ eight human annotators\footnote{Different from the annotators in Phase 1. Each annotator gets 247 sentences generated by PAC and WAC, each corresponding to the same set of parallel sentences.} to provide a rating between 1 (low quality) to 10 (high quality) to the generated Hinglish sentences. Table \ref{tab:guidelines} shows the annotation guidelines to rate the sentences. Figure \ref{fig:dataset_stat}a and \ref{fig:dataset_stat}b shows the distribution of the annotators' rating and their disagreement, respectively.

\begin{table*}[!tbh]
\resizebox{\hsize}{!}{
\begin{tabular}{|c|l|}
\hline \textbf{Task}            & \multicolumn{1}{c|}{\textbf{Guidelines}}                                        \\ \hline
\textbf{\begin{tabular}[c]{@{}c@{}}Hinglish text \\ generation\end{tabular}} & \begin{tabular}[c]{@{}l@{}}1. The Hinglish sentence should be written in Roman script.\\ 2. The Hinglish sentence should have words from both the source languages.\\ 3. Avoid using new words, wherever possible, that are not present in both sentences. \\ 4. If the source sentences are not the translation of each other, mark the sentence pair as ``\#''.\end{tabular} \\ \hline
\textbf{Quality rating}                                                      & \begin{tabular}[c]{@{}l@{}}The rating depends on the following three factors:\\ 1. The similarity between the generated Hinglish sentence and the source sentences.\\ 2. The readability of the generated sentence.\\ 3. The grammatical correctness of the generated sentence.\end{tabular}             \\ \hline
\end{tabular}}
\caption{Annotation guidelines to the annotators for the two different tasks.}
\label{tab:guidelines}
\end{table*}

\noindent \textbf{Data format}: Table \ref{tab:example} shows an instance from the dataset. In total, we have 3,952 instances\footnote{Two synthetic Hinglish sentences are generated for each parallel sentence pair.} in the dataset where each data instance \textit{i} for subtask-1 (see Section \ref{sec:subtask1}) is represented as \textbf{X1$_i$}=\{Eng$_i$, Hin$_i$, Synthetic\_Hing$_i$\} and \textbf{y1$_i$}=\{Average\_rating$_i$\}. For subtask-2 (see Section \ref{sec:subtask2}), the instance \textit{j} is represented as \textbf{X2$_j$}=\{Eng$_j$, Hin$_j$, Synthetic\_Hing$_j$\} and \textbf{y2$_j$}=\{Annotator\_disagreement$_j$\}. In addition, we provide at least two human generated Hinglish sentences corresponding to each data instance for both the subtasks. We shuffle and split the dataset in the ratio 70:10:20 with 2766, 395, and 791 data instances in train, validation, and test respectively. The more detailed description of the dataset is available in \cite{srivastava2021hinge}.

\section{The Two Tasks}
\label{sec: task}
\subsection{Subtask 1: Quality rating prediction}
\label{sec:subtask1}
The first subtask is predicting the quality rating of the code-mixed text. The participating teams can use the English, Hindi, and human-generated Hinglish sentences to predict the average rating\footnote{We take the greatest integer i $\leq$ average of the two rating scores.} as provided by the human annotators to the synthetic Hinglish sentences. In addition, we seek the teams to answer the following research questions implicitly with their experiments (not an exhaustive list):

\begin{itemize}[noitemsep,nolistsep,leftmargin=*]
  \item \textbf{RQ1.1}: Do the quality of source English and Hindi sentences impact Hinglish sentences' quality? 
  \item \textbf{RQ1.2}: Does the quality of Hinglish text generated by humans has any correlation with the quality of Hinglish text generated synthetically?
  \item \textbf{RQ1.3}: Does the dominance of a language (English or Hindi) present in the Hinglish sentence impact the rating provided by the humans?
  \item \textbf{RQ1.4}: How does the semantic and the syntactic correctness of the Hinglish sentence influence its quality?
\end{itemize}

\begin{table*}[t]
\centering
\includegraphics[width=1.0\linewidth]{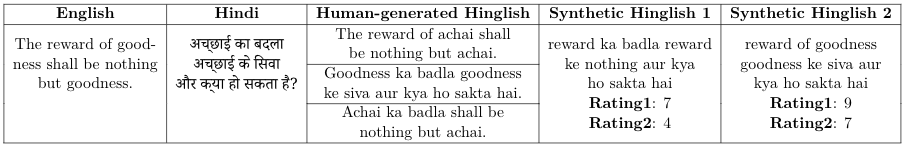}  
\caption{Example human-generated and synthetic Hinglish sentences from the dataset along with the source English and Hindi sentences. Two different human annotators rate the synthetic Hinglish sentences on the scale 1-10 (low-high quality).}
\label{tab:example}
\end{table*}


\subsection{Subtask 2: Annotators' disagreement prediction}
\label{sec:subtask2}
The next subtask is predicting the disagreement between the ratings provided by the human annotators to the synthetic Hinglish sentences. We calculate the disagreement between the ratings as the absolute difference between the two rating scores. Additionally, we seek the participating teams to answer the following research questions implicitly with their experiments (not an exhaustive list):
\begin{itemize}[noitemsep,nolistsep,leftmargin=*]
    \item \textbf{RQ2.1}: Does the quality of sentences in the source languages (English and Hindi) have any influence on the quality of the synthetic Hinglish sentences as seen by different individuals?
    \item \textbf{RQ2.2}: Does the quality of human-generated Hinglish sentence has any correlation with the quality of synthetic Hinglish text as seen by different individuals?
    \item \textbf{RQ2.3}: Do humans have a language bias while rating the quality of the code-mixed text?  
    \item \textbf{RQ2.4}: Do the similarity between human-generated and synthetic Hinglish sentences influence the annotators' disagreement?
\end{itemize}

\section{Evaluation}
We use the following three evaluation metrics:
\begin{itemize}[noitemsep,nolistsep,leftmargin=*]
    \item \textbf{F1-score (FS)}: We use the weighted F1-score to evaluate the system performance. The score ranges from 0 (worst) to 1 (best). 
    \item \textbf{Cohen's Kappa (CK)}: We use the Cohen's Kappa score to measure the agreement between the predicted and the actual rating. 
    The score ranges from $\le0$ (high disagreement) to $1$ (high agreement).
    \item \textbf{Mean Squared Error (MSE)}: MSE suggests the difference between the actual and the predicted scores. A low MSE score is preferred, with zero being the lowest possible score.
\end{itemize}
For the first subtask, we use all three metrics, whereas we use FS and MSE to evaluate the second subtask.
\section{Pilot Experiment}
We conducted a simple pilot experiment with a SOTA multilingual contextual language model M-BERT~\cite{devlin2019bert}. We fine-tune the pre-trained M-BERT model by adding one hidden-layer neural network on the top. We use the Relu activation function, AdamW optimizer with 0.03 learning rate, cross-entropy loss, and a batch size of 32. We use the contextual word-embedding corresponding to the synthetic Hinglish sentences in the dataset as an input to the model. The architecture remains the same for both subtasks. Table~\ref{tab:result} shows the result of the baseline experiment. We observe that the fine-tuned version of  M-BERT performs poorly on both the subtasks on all the evaluation metrics. These language models are not as effective for both the subtasks as compared to other code-mixed text classification tasks where they seem to perform better than other rule-based and neural approaches \cite{gupta2021task,winata2021multilingual}. Overall, we observe the poor performance of M-BERT based classifier on the current two subtasks. Specifically, for subtask 1, the agreement (measured by CK score) between predicted rating and human rating is close to 0. These results present an excellent opportunity to propose a shared task that enhances the generated code-mixed text quality estimation.

\begin{table}[t]
\small{
\resizebox{\hsize}{!}{
\begin{tabular}{|c|c|c|c|c|c|}
\hline
\multirow{2}{*}{} & \multicolumn{3}{c|}{\textbf{Subtask 1}}  & \multicolumn{2}{c|}{\textbf{Subtask 2}}  \\ \cline{2-6} 
                  & \textbf{FS} & \textbf{CK} & \textbf{MSE} & \textbf{FS}  & \textbf{MSE} \\ \hline
\textbf{Val}      &       0.202      &  0.003           &      2.797        &     0.209               &       4.987       \\ \hline
\textbf{Test}     &     0.256        &     0.092        &    2.628          &     0.242               &        4.317      \\ \hline
\end{tabular}}}
\caption{Evaluation of the pilot experiment.}
\label{tab:result}
\end{table}

\section{Conclusion}
In contrast to the non-code-mixed text, the noisy and informal nature (e.g., spelling variation, missing punctuation, and language switching) of the code-mixed text makes the quality evaluation more loosely defined. Consequently, we need to build models that can effectively gauge the human perception of the quality of the code-mixed text. This shared task will help to build efficient and robust code-mixed text generation and evaluation systems. 

\bibliographystyle{acl_natbib}
\bibliography{anthology,acl2021}


\flushend
\end{document}